\pdfoutput=1

\documentclass[11pt]{article}

\usepackage[preprint]{acl}

\usepackage{times}
\usepackage{latexsym}

\usepackage[T1]{fontenc}

\usepackage[utf8]{inputenc}

\usepackage{microtype}

\usepackage{inconsolata}

\usepackage{graphicx}

\usepackage{hyperref}       
\usepackage{url}            
\usepackage{booktabs}       
\usepackage{amsfonts}       
\usepackage{nicefrac}       
\usepackage{xcolor}         
\usepackage{multirow}
\usepackage{comment}
\usepackage{wrapfig}
\usepackage{tablefootnote}
\usepackage{caption}
\usepackage{subcaption}
\usepackage{colortbl}
\usepackage{amsmath}

\usepackage{breakurl}
\usepackage[breaklinks]{hyperref}
\usepackage[normalem]{ulem}
\useunder{\uline}{\ul}{}
\usepackage[textsize=scriptsize]{todonotes}
\usepackage{adjustbox} 

\title{Towards Robust Knowledge Representations in Multilingual LLMs for \texttt{Equivalence} and \texttt{Inheritance} based Consistent Reasoning}

\author{Gaurav Arora \\
  Amazon \\
  \texttt{gaurvar@amazon.com} \\\And
  Srujana Merugu \\
  Amazon \\
  \texttt{smerugu@amazon.com} \\
  \AND
  Shreya Jain \\
  IIT Jammu\thanks{Contributed to this work during her internship at Amazon} \\
  \texttt{2020uee0135@iitjammu.ac.in} \\\And
  Vaibhav Saxena \\
  Amazon \\
  \texttt{saxenvai@amazon.com} \\
  }

\begin{document}
\maketitle
\begin{abstract}
Reasoning and linguistic skills form the cornerstone of human intelligence, facilitating 
problem-solving and decision-making. Recent advances in Large Language Models (LLMs) have led to impressive linguistic capabilities and emergent reasoning behaviors, fueling widespread adoption across 
application domains.
However, LLMs still struggle 
with complex reasoning tasks, 
highlighting their systemic limitations.
In this work, we focus on evaluating whether LLMs have the requisite representations to reason using two foundational relationships: "equivalence" and "inheritance". 
We introduce novel tasks and 
benchmarks spanning six languages and observe 
that current SOTA LLMs often produce conflicting answers to the same questions across languages in 17.3-57.5\% of cases and violate inheritance constraints in up to 37.2\% cases. To enhance 
consistency 
across languages, we propose novel  "Compositional Representations" where tokens  are represented as
composition of equivalent tokens across languages, with resulting 
conflict reduction (up to -4.7\%)
indicating
benefits of shared LLM representations.

\end{abstract}

\section{Introduction}
\label{sec:intro}

Reasoning is the capacity to employ logic and analyze relationships among entities to extrapolate from known evidence to derive new insights. Language significantly bolsters this process by supplying the necessary structure and vocabulary  for encoding complex ideas, thus facilitating hypothesis generation and evaluation.  The intricate connection between  linguistic and reasoning capabilities is a hallmark of human intelligence, enabling abstract thinking, problem-solving, and  decision-making.

Recent advancements in LLMs such as ChatGPT \cite{chatgpt_website} and Claude \cite{claude_website} 
showcase 
their exceptional language generation capabilities and their potential to boost performance across diverse Natural Language Processing (NLP) tasks \cite{ahuja2023mega}, with multiple studies also pointing to emergent reasoning abilities at scale  \cite{wei2022emergent}. However, LLMs continue to face challenges with complex reasoning tasks such as planning and 
problem-solving, 
indicating that
their expansive modeling capacity and extensive training regime might enable them to mask deeper systemic shortcomings through superficial reasoning. As LLMs increasingly permeate 
applications
catering to multilingual 
users with complex needs,
gaining a deeper understanding of their functioning becomes imperative, as it could uncover systemic gaps and pave the way for superior models.


\begin{figure}
  \includegraphics[width=\columnwidth]{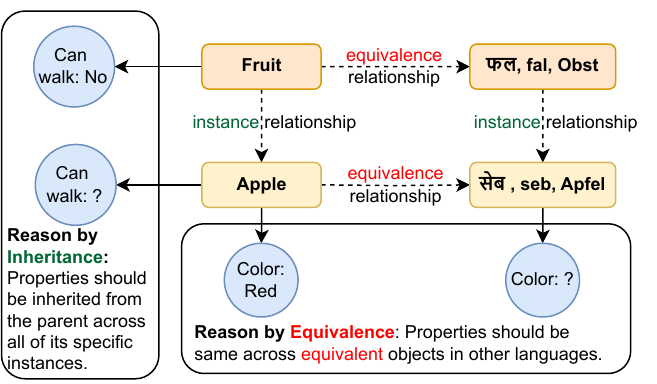}
  \caption{An example of Reasoning by "Equivalence" and Reasoning by "Inheritance" based on \emph{existence} of equivalence/inheritance relationship between concepts.}
  \label{fig:intro_example}
  \vspace{-0.15in}
\end{figure}

\begin{figure*}
    \centering
    \includegraphics[width=\textwidth]{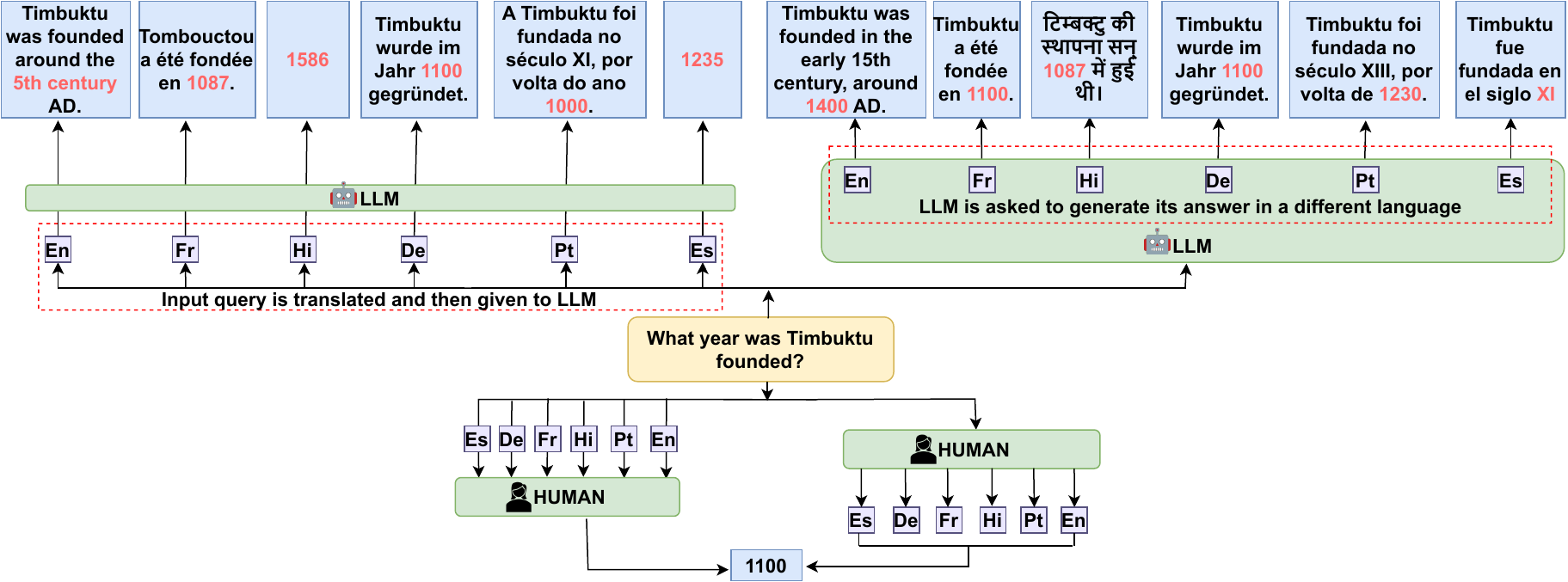}
    \caption{An example of LLM (Claude-v1 Instant) lacking "equivalence relationship between equivalent concepts across languages" due to tightly coupling of knowledge representation and language expression unlike in Humans.}
    \label{fig:rq1_figure}
    \vspace{-0.15in}
\end{figure*}



Robust reasoning hinges on the availability of powerful 
constructs such as entity-relation (ER) graphs  and rules to interpret  relationships.
For instance, an ER graph with entities A, B, and C, where "A is the father of B" and "C is the wife of A" allows us to deduce that C is likely the mother of B, based on interpretation of the relationships "father" and "wife". 
While there are myriad relationships underpinning reasoning such as “cause and effect“ and ”comparison“,  in our current work, we focus on "equivalence" and "inheritance" due to their predominance in enhancing the efficiency of logical inference through property transfer, which is also reflected in their adoption as  core constructs of knowledge representation and programming languages \cite{minsky1974framework}.
Fig \ref{fig:intro_example} illustrates these relationships showing how humans create the necessary representations
of "\texttt{Apple}" to reason across equivalent objects ("\texttt{seb}" : "\texttt{Apple}" in Hindi (transliterated), "\texttt{Apfel}" : "\texttt{Apple}" in German, etc) independent of the language/script of expression \underline{(Reasoning by Equivalence)}, and inherit properties from the abstract concept "Fruit" across all of its specific instances ("\texttt{Apple}", "\texttt{Orange}" etc) (\underline{Reasoning by Inheritance}). Here, we expect representations of equivalent objects to be similar, while that of inherited objects satisfy transitivity. These representations are crucial for efficient learning, knowledge sharing, and updation of beliefs.

Typically, humans create abstractions using denotational semantics, i.e., a word's meaning is defined by objects it describes, 
which is the favored approach in logical theory.
In contrast, LLMs use distributional semantics, i.e., a word's meaning stems from  the training data context, which can be problematic when the data has gaps, such as infrequent connections between equivalent words across languages.
Hence, despite impressive  performance on NLP tasks\cite{ahuja2023mega},
it is unclear if LLMs create the necessary representations
within and across languages to support consistent reasoning across 
equivalent and inherited objects.


\noindent\textbf{Contributions.}
In this work, we focus on whether LLMs have the requisite representations to 
reason by equivalence and inheritance across 
languages and make the below contributions. \\
\noindent 1. We introduce a novel task and parallel benchmark datasets of factoid QA to evaluate "Reasoning by Equivalence" in LLMs and assess the  performance of multiple 
SOTA LLMs  on this task across 6 languages (English, French, Spanish, German, Portuguese and Hindi). On our benchmarks,  LLMs generate conflicting answers across languages in 17.3-57.5\% of cases indicating a significant gap. We also perform a controlled experiment to identify factors promoting consistency across languages 
and find a strong positive correlation with similarity in script and typology.
 
\noindent 2. We present another task and associated new benchmark  to evaluate "Reasoning by Inheritance" in LLMs and study the proficiency of multiple SOTA LLMs  across six languages. 
Our results indicate that LLM answers 
violate inheritance constraints in up to 37.2\%  cases across these languages 
with most violations observed in Hindi.

\noindent 3. We also propose a novel method for constructing "Compositional Representations" in LLMs by representing tokens  as  composition of other equivalent tokens in vocabulary, which grants the model access to (otherwise) distant representations of equivalent objects across languages, 
thereby facilitating improved knowledge sharing and reduction in conflicts with gains up to 4.7\% compared to baselines.

To the best of our knowledge, this is the first quantitative study 
of LLM reasoning via equivalence and inheritance across languages. We intend to share the benchmarks as a community resource and to ensure reproducibility. Note that even when the desired equivalence and inheritance relationships hold and properties transfer correctly (our current focus), there may be gaps in LLM's  multi-step reasoning process due to other factors as we discuss in detail in Appendix \ref{appendix:existence_vs_using}.

\section{Related Work}
\label{sec:related_work}
\textbf{Multilingual NLP.} LLMs like ChatGPT \cite{chatgpt_website}, GPT-4 \cite{openai2024gpt4}, Claude \cite{claude_website}, BLOOMZ \cite{muennighoff2023crosslingual}, XGLM \cite{lin-etal-2022-shot} have shown impressive performance on standard multilingual NLP tasks and benchmarks \cite{ahuja2023mega, zhao2023survey, enis2024llm, ahuja2024megaverse}. Despite extensive evaluations and the existence of parallel multilingual datasets such as MLQA and XQUAD \cite{ahuja2024megaverse}, to the best of our knowledge, there is  no prior work or tailored benchmarks for assessing LLMs' ability to reason by equivalence and inheritance across multiple languages. Further,  there is a chance of public benchmarks with  duplicated knowledge across languages being included in LLM training data, rendering reasoning related assessments  unreliable. Our study is the first to create controlled benchmarks and evaluate LLMs on these reasoning tasks to  identify gaps and potential contributing factors.
 
\textbf{Reasoning in LLMs.} Reasoning abilities of LLMs have been studied for problem solving, decision making, and critical thinking \cite{huang2023reasoning, wei2022emergent, bubeck2023sparks}. Prior work has also looked at evaluating ability of encoder only models like BERT \cite{bert_paper},
 RoBERTa \cite{roberta_paper} to understand ontological knowledge \cite{wu-etal-2023-plms}. In this work, we focus on reasoning based on two foundational relationships: equivalence and inheritance and evaluate popular LLMs on these dimensions across multiple languages. In recent years, there  have been advances \cite{embed2sym, lazzari2024sandra, marconato2023neurosymbolic} in neuro-symbolic architectures that combine symbolic and sub-symbolic components to enable efficient computation of symbolic representations and deductive reasoning. However, these works do not present a detailed analysis of representations of equivalent or related entities and these methods also entail much higher computational costs, limiting their adoption. 

\textbf{Representation Learning in NLP.} Improving distributed representations led to significant performance improvements in past \cite{representation_learning_nlp, bert_paper, mikolov2013efficient}. 
Prior work on adapting attention mechanisms to bridge gaps across disparate but related inputs such as  translated/transliterated data 
has led to improved multilingual representations  \cite{conneau-etal-2020-unsupervised, khanuja2021muril, arora-etal-2023-comix}.
In our current work, we 
bridge the gap between distant representation spaces of  various languages by adapting the attention mechanism to  better utilize the token-language mapping.

\section{Reasoning by Equivalence}
\label{sec:rq1}
Reasoning by equivalence is a  core building block that enables efficient and scalable reasoning across contexts, with the efficiency being determined by the size of the equivalence classes (sets of equivalent objects). 
Construction of these “equivalence classes”, i.e., "abstract concepts"  from specific contexts
and reusing these abstract concepts flexibly beyond the specific contexts \cite{an2023does, Mitchell_2021, kumar2023disentangling, GIUNCHIGLIA1992323, hull1920quantitative}, is a natural human skill. 
The human ability to acquire knowledge from one language (e.g., “apple is red”)
and construct representations shared across languages  as in Fig \ref{fig:intro_example} is a prime example.  Similar to multilingual humans, LLMs also see large 
amount of multilingual data during pre-training 
\cite{blevins-zettlemoyer-2022-language}. 
For instance, pretraining data of GPT-3 and BLOOM spanned 119 and 46 languages \cite{gpt3_paper, bloom_paper} respectively.
In this section, we evaluate if SOTA LLMs also have the ability to reason by equivalence across languages given their impressive multilingual capabilities, and if this ability 
is due to shared representations (i.e. existence of equivalence relationship) or duplication of knowledge across languages. 
Note that the existence of "equivalence relationships encoded in LLM representations"
is a fundamental prerequisite for complex reasoning even though it does not guarantee
that LLMs can effectively leverage it for multi-step logical reasoning due to gaps in LLM’s inference mechanism based on associative attention (see Appendix \ref{appendix:existence_vs_using}). 

\begin{table*}
\centering
\begin{tabular}{lcccccc}
\hline
\textbf{}                  & \textbf{En-Fr} & \textbf{En-Es} & \textbf{En-De} & \textbf{En-Pt} & \textbf{En-Hi} & \textbf{Avg}   \\ \hline
\textbf{Claude v1 Instant} & 31.41          & 32.66          & 30.81          & 31.99          & 56.31          & 36.64          \\
\textbf{Claude v2}         & 22.20          & 22.54          & 21.75          & 22.89          & 47.02          & 27.28          \\
\textbf{Claude v3 Sonnet}  & \textbf{20.99}          & \textbf{21.23}          & \textbf{20.60}          & \textbf{21.49}          & \textbf{43.51} & \textbf{25.56} \\
\textbf{BLOOMZ-7B}         & 37.23          & 34.67          & 50.88          & 33.95          & 52.69          & 41.88          \\
\textbf{XGLM-7.5B}         & 42.99          & 42.49          & 38.74          & 41.74          & 57.53          & 44.70          \\ \hline
\end{tabular}
\caption{Conflict rate between LLM answers to parallel En-XX questions, XX = {[}Fr, Es, De, Pt, Hi{]}.}
\label{tab:rq1_res}
\end{table*}

\begin{table*}
\centering
\begin{tabular}{lcccccc}
\hline
\textbf{}                  & \textbf{En-Fr} & \textbf{En-Es} & \textbf{En-De} & \textbf{En-Pt} & \textbf{En-Hi} & \textbf{Avg}   \\ \hline
\textbf{Claude v1 Instant} & 29.91          & 30.44          & 29.72          & 30.28          & 42.92          & 32.65          \\
\textbf{Claude v2}         & 20.45          & 21.13          & 20.52          & 21.38          & \textbf{28.14} & 22.32          \\
\textbf{Claude v3 Sonnet}  & \textbf{17.89} & \textbf{18.06} & \textbf{17.34} & \textbf{17.99} & 28.61          & \textbf{19.97} \\ \hline
\end{tabular}
\caption{Conflict rate between LLM responses in {[}En, Fr, Es, De, Pt, Hi{]} for input questions in En.}
\label{tab:rq1_res2}
\end{table*}

\subsection{How good are LLMs at exhibiting “Reasoning by Equivalence”?}

We evaluate if LLMs exhibit “Reasoning by Equivalence” specifically across languages by estimating
how dependent LLM's answers are on the language of input/output expression.  
A high dependency indicates  tight coupling of the knowledge representation with the language and points to lack of shared abstractions and limited ability to reason by equivalence with knowledge duplication.  Fig \ref{fig:rq1_figure} shows an example where Claude-v1 instant's answers are significantly dependent on the language of input/output expression. Below, we  outline our methodology, dataset, metrics and results.

\subsubsection{Methodology}
Let $R=\{r_1,\cdots r_L\}$ be a set of languages with $[X^{r_1}X^{r_2}, ... X^{r_L}]$ denoting parallel questions across $r_i \in R$. We perform two assessments.


\noindent \textbf{D}ependency on \textbf{I}nput \textbf{L}anguage \textbf{(DIL)}. Given $L$ parallel questions $[X^{r_1},X^{r_2}, ... X^{r_L}]$; we generate LLM answers\footnote{Temperature=0 across the paper for deterministic outputs.} $Y^{r_i}$ for each $X^{r_i}$ independently. Choosing an anchor language $r_\mathrm{anchor} \in R$, we check
for each $r \in R \setminus \{r_\mathrm{anchor}\}$ if $Y^{r_\mathrm{anchor}}$ and $Y^{r}$ conflict with each other .

 
\noindent \textbf{D}ependency on \textbf{O}utput \textbf{L}anguage \textbf{(DOL)}.
Given $r_\mathrm{anchor}$, we input  $X^{r_\mathrm{anchor}}$ into the LLM and generate answers $Y^{r}$ for $X^{r_\mathrm{anchor}}$ for all $r \in R$ independently. Then, we check for each $r \in R \setminus \{r_\mathrm{anchor}\}$ if $Y^{r_\mathrm{anchor}}$ and $Y^{r}$ conflict with each other.

Here, two answers are called conflicting if they contain contradictory information and not merely if there are different or one of them is non-informative. For our experiments, we consider English (En), French (Fr), Spanish (Es), German (De), Portuguese (Pt) and Hindi (Hi) languages, i.e. $R= \{\mathrm{En}, \mathrm{Fr}, \mathrm{Es}, \mathrm{De}, \mathrm{Pt}, \mathrm{Hi}\}$ and $r_\mathrm{anchor} = \mathrm{En}$.
Since authors in \cite{lin-etal-2022-shot,ahuja2023mega} show that English instructions in the prompt perform better than instructions written in the native language for non-English languages, we tune the English instructions in prompt separately for each LLM and then keep 
these
consistent for that LLM across all DIL and DOL experiments. 

\subsubsection{Dataset and Metrics}
\label{sec:reasoning_equivalence_data}
\textbf{Dataset.} To ensure feasibility of automated evaluation via LLM-based judges and reduce variations due to subjective interpretation and cultural variations, our evaluation focused primarily on objective factual/attribute-based questions on entities. We prepare $\mathrm{En}$ factual questions dataset consisting of 88,334 questions on well known named entities and translate the dataset to $\mathrm{Fr}$, $\mathrm{Es}$, $\mathrm{De}$, $\mathrm{Pt}$ and $\mathrm{Hi}$ using AWS Translate \cite{aws_translate}. See Appendix \ref{appendix:rq1_data_prep} for more details on the dataset and Fig \ref{fig:rq1_sample_data} for a few sample questions.\\
\textbf{Metrics.} We compute the conflicts among answers generated by LLM for different input/output expression languages.\footnote{Since we care about consistent and common knowledge representation in LLMs for equivalent concepts, we only assess conflicting LLM responses to equivalent questions and not worry about factual accuracy of responses.}
We define conflict rate between $(r_i, r_j)$ language pair as fraction of total answer pairs which are conflicting, i.e. 
    $\mathrm{ConflictRate}(r_i, r_j) = \frac{\Sigma_{k=1}^{|D|}\mathrm{J}(Y_k^{r_i}, Y_k^{r_j})}{|D|} $, where $|D|$ is dataset size and $J$ returns $1$ if $(Y_k^{r_i}, Y_k^{r_j})$ are conflicting, else returns $0$. We use Claude v3 Sonnet as the judge $J$ with prompt shown in Fig \ref{fig:judge_prompt} in Appendix \ref{appendix:judge_precision}. Table \ref{tab:judge_prec} in Appendix \ref{appendix:judge_precision} shows that the average precision of our judge is >95\%.

\begin{figure*}
\centering
\adjustbox{max width=.97\textwidth}{
\begin{subfigure}{.5\linewidth}
  \centering
  \includegraphics[width=0.8\linewidth,height=5cm]{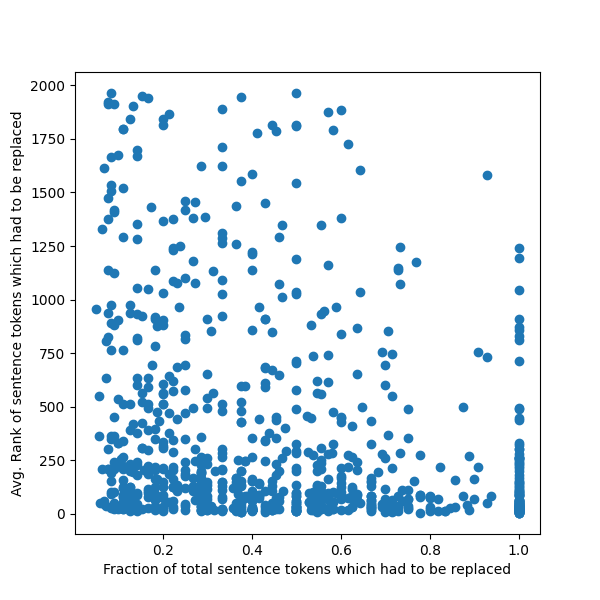}
  \caption{German}
  \label{fig:en_de_controlled_scatter}
\end{subfigure}%
\begin{subfigure}{.5\linewidth}
  \centering
  \includegraphics[width=0.8\linewidth,height=5cm]{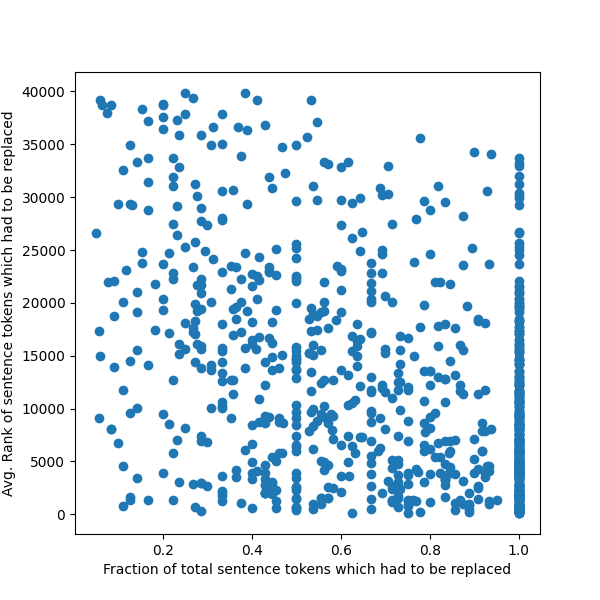}
  \caption{Hindi}
  \label{fig:en_hi_controlled_scatter}
\end{subfigure}}
\caption{Avg. Rank and Fraction of tokens which had to be replaced in German and Hindi with parallel English token to achieve consistent answer as English.}
\label{fig:controlled_scatter}
\vspace{-0.15in}
\end{figure*}

\subsubsection{Analysis and Results}

Table \ref{tab:rq1_res} shows conflict rate for various LLMs for DIL task.\footnote{GPT-3.5 had similar results as Claude v3 Sonnet but we could not add those results due to organization policy.\label{gpt35_forbid_footnote}} We can see that conflict rate reduces 
with increase in model strength.
Open source models lag behind closed source models by a significant margin with 25-44\% average conflict rate across different LLMs and languages. Since we establish from results in Table \ref{tab:rq1_res} that LLMs are highly dependent upon input expression language, and that this dependency is consistent across varied open-source and closed-source LLMs of varied size, we evaluate only Claude family models for DOL task as shown in Table \ref{tab:rq1_res2}. 
We observe similar trends as average conflict rate of 19-32\%. Both these results show that knowledge representation is tightly coupled with expression/language in LLMs, indicating a lack of right abstractions and limited knowledge sharing across languages in LLMs. Fig \ref{fig:rq1_DIL_sample_errors} and Fig \ref{fig:rq1_DOL_sample_errors} in Appendix \ref{appendix:rq1_analysis} show sample conflicting answers from various LLMs to equivalent questions from DIL and DOL tasks respectively.

\subsection{Factors affecting LLMs ability to exhibit “Reasoning by Equivalence” }
\label{sec:controlled_exp}

To better understand  knowledge transfer and source of conflicts across languages in LLMs, we perform a controlled experiment wherein we create synthetic QnA data  with non-existent named entities that LLM does not have any prior knowledge on
and train it on synthetic data in one language and test for its transfer in other languages.

\subsubsection{Controlled Experiment}
\label{sec:controlled_experiment_setup}

\textbf{Dataset.} We create synthetic data of non-existent named entities and hallucinated articles about those  entities using Claude v1-instant. We also generate factual questions about synthetic named entities which can be answered only from hallucinated articles. We only keep those questions for which Claude's answers with and without the article conflict with each other to ensure any LLM is unlikely to have any prior knowledge of our synthetic data. Our synthetic QnA dataset has 2063 synthetic named entities and has 32016 QnA pairs. 
As we perform the  controlled experiment with $\mathrm{En}, \mathrm{De}, \mathrm{Hi}$ and $\mathrm{HiEn}$ (transliterated Hindi), and our synthetic question set is in $\mathrm{En}$, we also translate it to $\mathrm{De}$ and $\mathrm{Hi}$ using AWS Translate, and transliterate to $\mathrm{HiEn}$ using IndicTrans \cite{Bhat:2014:ISS:2824864.2824872}.

\textbf{Experiment Setup.} We train XGLM-4.5B on unique 25\% of the synthetic data for each language and test on its parallel data in other three languages. Specifically, we train on concatenated $D^{\mathrm{en}}_{1}, D^{\mathrm{hi}}_{2}, D^{\mathrm{de}}_{3}, D^{\mathrm{hien}}_{4}$ data where $D^{r}_{s}$ denotes synthetic data in $r^{th}$ language from $s^{th}$ quarter of $D^{r}$. For $s=1$, $\mathrm{En}$ is the anchor language and we evaluate knowledge transfer to $\mathrm{Hi}, \mathrm{De}, \mathrm{HiEn}$ by looking at conflicts between LLM answer to $\mathrm{En}$ question from $D^{\mathrm{en}}_{1}$ and LLM answer to parallel $\mathrm{Hi}, \mathrm{De}, \mathrm{HiEn}$ questions from $D^{\mathrm{hi}}_{1}, D^{\mathrm{de}}_{1}, D^{\mathrm{hien}}_{1}$ respectively. As shown in Fig \ref{fig:controlled_exp_setup_figure} in Appendix \ref{appendix:controlled_exp}, same procedure is followed for $s \in \{2,3,4\}$ where $\mathrm{Hi}, \mathrm{De}, \mathrm{HiEn}$ are anchor languages respectively.


\begin{figure}
  \includegraphics[width=\columnwidth]{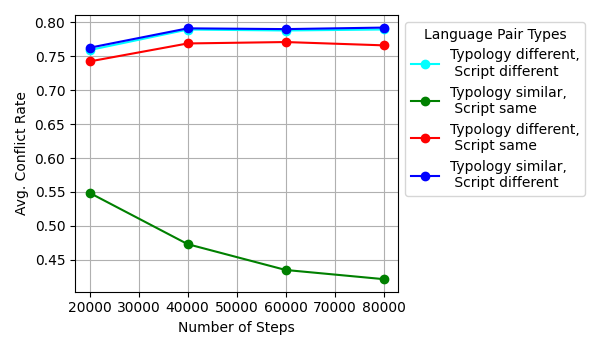}
  \caption{Conflict rate for different language pair types.}
  \label{conflict_rate_vs_steps}
  \vspace{-0.15in}
\end{figure}

\subsubsection{Results and Analysis}
\label{sec:controlled_exp_res}

Fig \ref{conflict_rate_vs_steps} shows there is significantly higher knowledge transfer between languages with similar typology and same script, as compared to 
the 
pairs where either typology or script is different (see Fig \ref{fig:conflict_rate_with_steps} in Appendix \ref{appendix:controlled_exp} for conflict rate of all language pairs individually). For instance, conflict rate for En-De is much lower than that of En-HiEn, which in turn is lower than that of En-Hi.



We define \textbf{Rank} of a token $x$ in a non-anchor question as 
the \textit{position} 
at which its parallel token from anchor question occurs if we sort all tokens in the vocabulary by cosine similarity with $x$ in descending order. Rank of a token captures the relative 
proximity to the anchor language's parallel token in the embedding 
space.
To assess what it would take to get a non-conflicting answer, we replace tokens in the non-anchor question with their parallel anchor question tokens in the descending order of rank, i.e., farthest (non-anchor,anchor) parallel tokens are replaced first, till we obtain a non-conflicting answer.  Fig \ref{fig:en_de_controlled_scatter} shows dense region of $\mathrm{De}$ tokens with small ranks which had to be replaced by their parallel anchor question tokens to reach a non-conflicting consistent answer. This shows that 
having close enough representations for equivalent tokens in different languages might also not be enough, and they have to be same for LLM to learn consistent knowledge. This limitation stems from the representation space LLMs operate in since we project discrete symbols in the continuous embedding space.  On the other hand, for $\mathrm{Hi}$ in Fig \ref{fig:en_hi_controlled_scatter}, average rank of replaced tokens is significantly higher relative to $\mathrm{De}$, likely due to the  typology and script differences in case of $\mathrm{En-Hi}$,
which points to the need for 
near similar representations to share knowledge.

These results show that consistency of LLMs for distant languages is likely to stem from duplication of knowledge across languages in the training data, whereas for languages with similar typology and script,  knowledge sharing occurs due to similar representation of equivalent tokens and information propagation. See Appendix \ref{appendix:effect_of_duplication} for more discussion on the effect of duplication of knowledge across languages in LLM's training data.




\section{Reasoning by Inheritance}
\label{sec:rq2}
\begin{table*}
\centering
\begin{tabular}{lccccccc}
\hline
\textbf{}                  & \textbf{En}   & \textbf{Fr}   & \textbf{Es}   & \textbf{De}   & \textbf{Pt}   & \textbf{Hi}    & \textbf{Avg}  \\ \hline
\textbf{Claude v1 Instant} & 3.36          & 18.33         & 17.16         & 9.12          & 15.05         & 36.5           & 16.59         \\
\textbf{Claude v2}         & 3.37          & 4.55          & 11.69         & 4.96          & 6.78          & 18.53          & 8.31          \\
\textbf{Claude v3 Sonnet}  & \textbf{0.13} & \textbf{4.43} & \textbf{7.27}          & \textbf{1.86} & \textbf{5.41} & \textbf{14.69} & \textbf{5.63} \\
\textbf{BLOOMZ-7B}         & 4.78          & 8.76          & 7.49          & 8.08          & 10.4          & 23.27          & 10.46         \\
\textbf{XGLM-7.5B}         & 36.61         & 30.87         & 35.65         & 36.5          & 37.27         & 33.59          & 35.08         \\ \hline
\end{tabular}
\caption{Conflict rate of LLM answers on inheritance-based questions on common concepts in En, Fr, Es, De, Pt, Hi.}
\label{tab:rq2_res}
\end{table*}

Reasoning by Inheritance is also a key building block of common sense and logical reasoning in humans enabled by concept abstractions.
Humans identify common patterns amongst instances of the same type and create abstract concepts to reason consistently across all specific instances by inheriting properties from the abstract concept as shown in Fig \ref{fig:intro_example}. In this section, we investigate if SOTA LLMs can use their ontological knowledge and inherit properties from abstract concepts consistently across various specific instances of the abstract concept within multiple languages. 

\begin{figure}
\centering
\includegraphics[width=\columnwidth]{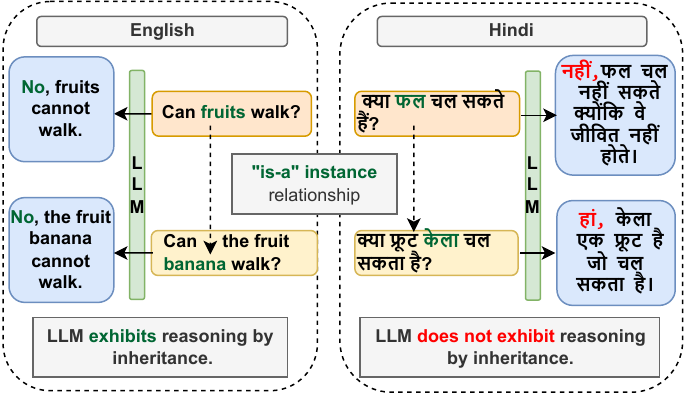}
\caption{An example of LLM (Claude v1 Instant) exhibiting reasoning by inheritance in $\mathrm{En}$ but not in $\mathrm{Hi}$.}
\label{fig:rq2_figure}
\vspace{-0.15in}
\end{figure}


We evaluate if an LLM exhibits “Reasoning by Inheritance” within a language by checking if specific instances of an abstract parent concept inherit properties of the parent without conflicts.
Fig \ref{fig:rq2_figure} shows an example wherein Claude-v1 instant does not exhibit reasoning by inheritance in $\mathrm{Hi}$ due to the lack of the right abstractions.

\textbf{Dataset.} We prepare a set of 35 abstract concepts and 2396 well known named entities which are specific instances of those abstract concepts. For each one of the abstract concept, we hand-curate set of properties that all of its specific instances should inherit. We create templatized questions from those properties to prepare $\mathrm{En}$ dataset and translate it to $\mathrm{Fr}$, $\mathrm{Es}$, $\mathrm{De}$, $\mathrm{Pt}$ and $\mathrm{Hi}$ using AWS Translate \cite{aws_translate}. Fig \ref{fig:rq2_sample_data} in Appendix \ref{appendix:rq2_dataset} shows a sample of our dataset.

\textbf{Methodology.} We  evaluate the consistency of LLMs in inheriting and applying ontological knowledge consistently across specific instances of abstract concepts by directly asking in native language if the specific instance has the property of abstract concept and checking if the answer for abstract concept and specific instance are conflicting.

\textbf{Metrics.} We compute conflict rate as in the Section \ref{sec:rq1} but focus on  comparing LLM answers on an inheritable property for an abstract parent concept and that of its specific instances.

\textbf{Results.} Table \ref{tab:rq2_res} shows conflict rate for questions requiring reasoning by inheritance across six languages.\footref{gpt35_forbid_footnote} We notice that conflict rate is low for $\mathrm{En}$ for most models but quite high for $\mathrm{Hi}$, which likely has low representation in the model training corpus. This indicates that the ability to reason by inheritance likely depends on the amount of specific language data during training LLMs, which in turn points to gaps in  inductive biases in LLMs. Fig \ref{fig:rq2_sample_errors} in Appendix \ref{appendix:rq2_sample_errors} shows few LLM responses that violate inheritance constraints with conflicting answers for parent and child concepts.


\section{Compositional Representation (CoRe)}
\label{sec:comp_rep}

\begin{figure}
  \includegraphics[width=\columnwidth]{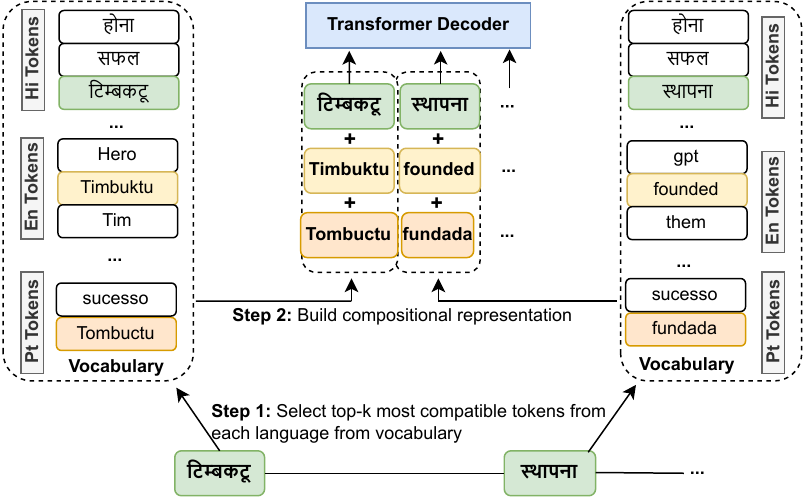}
  \caption{CoRe Illustration: LLM can access distant equivalent representations to permit knowledge sharing.}
  \label{fig:comp_rep_b}
  \vspace{-0.15in}
\end{figure}

We now consider mechanisms to mitigate lack of knowledge consistency across languages which emerged as a problem in the prior sections. Our analysis points to two key observations:

\noindent 1. “Identical” representations for equivalent concepts ensures perfect knowledge transfer while 
“distant” representations lead to separate copies of knowledge being learned.

\noindent 2. Languages from different families (such as  En and Hi) have  distant LLM representations for equivalent concepts resulting in low  knowledge sharing between them and high inconsistency unless duplicate information is fed in both languages.

Based on these observations, we propose CoRe with the aim of  bridging  distant representation spaces to enable greater knowledge sharing.  It hinges on the key idea that  representing a concept via composition of representations of all equivalent concepts across languages would enable LLM to maintain consistent knowledge for that concept across languages as shown in Fig. \ref{fig:comp_rep_b}.  This is in contrast to the current LLM models (Fig. \ref{fig:comp_rep_a}) where a sentence  input into the transformer decoder in LLMs does not have access to representations of equivalent tokens from distant languages and can only utilize the localised knowledge in that language’s representation space.

\begin{table*}[]
\centering
\resizebox{\textwidth}{!}{%
\begin{tabular}{cccccccccccccc}
\hline
\textbf{}                                                                    & \textbf{n} & \textbf{en-hi} & \textbf{en-de} & \textbf{en-hien} & \textbf{hi-en} & \textbf{hi-de} & \textbf{hi-hien} & \textbf{de-en} & \textbf{de-hi} & \textbf{de-hien} & \textbf{hien-en} & \textbf{hien-hi} & \textbf{hien-de} \\ \hline
XGLM-4.5B                                                                    & -          & 74.4           & 58.6           & 73.2             & 72.6           & 73.6           & 72.4             & 57.1           & 75             & 74.3             & 70.5             & 72.1             & 71.8             \\
\hline
\multirow{3}{*}{\begin{tabular}[c]{@{}c@{}}XGLM-4.5B\\  + CoRe\end{tabular}} & 5          & 73.5           & 55.6           & \textbf{71.9}    & 71.9           & 72.7           & 71.9             & 55.7           & 74             & \textbf{73.3}    & \textbf{68.4}    & 70.4             & 69.5             \\
                                                                             & 10         & \textbf{72.8}  & \textbf{53.9}  & 73.1             & \textbf{71}    & \textbf{72.2}  & 72.2             & \textbf{53.8}  & \textbf{73.9}  & 73.9             & 69.8             & \textbf{70}      & \textbf{69.1}    \\
                                                                             & 15         & 73.2           & 56.8           & 72.3             & 71.3           & 73.3           & \textbf{71.8}    & 56.8           & 75.5           & 73.5             & 68.7             & 71               & 69.2             \\ \hline
\end{tabular}%
}
\caption{Effect of CoRe on conflict rate for different language pairs.}
\label{tab:core_res}
\end{table*}

\subsection{Methodology}
\label{sec:core_methodology}

Attention \cite{Vaswani2017AttentionIA} is an essential mechanism of transformer architecture that converts an input sequence into a latent encoding  using  representational vectors formed from the input, i.e.,  queries, keys and values to determine the importance of each portion of input while decoding. Typically, in transformers, each token pays attention to other nearby tokens in the input sequence. Since this process can miss out on equivalent tokens in other languages, especially in the absence of a parallel multi-lingual corpus, we modify the learning approach to consider all tokens in the vocabulary as candidates for attention.  Specifically 
our methodology consists of two steps: (a) proximal token selection, and (b) construction of compositional representation, which we describe below.

\noindent \textbf{Step 1: Proximal token selection.} For a given token, we first select the top-$n$ proximal tokens across each  language based on compatibility of the  existing representations.  Formally, let  $X=[x_i]_{i=1}^N $ and $Z=[z_i]_{i=1}^N$ denote the sequence of input tokens and the associated embedding representations of size $d_o$. Let $U=[u_j]_{j=1}^M$ be all the vocabulary tokens  and $B=[b_j]_{j=1}^M$ be the associated embedding representations of size $d_o$. Further, let   $U^r$ be the vocabulary tokens associated with the language $r \in R$, where $R$ denotes the entire set of languages being considered. Let $Q=[q_i]_{i=1}^N$, $K=[k_j]_{j=1}^M$ and $V=[v_j]_{j=1}^M$  be the  sequences of query, key and value vectors of dimensions $d_{k}, d_{k}$ and $d_{v}$ respectively, given by  $q_i=z_iW_Q$, $k_j=b_jW_K$ and $v_j=b_jW_V$  where $W_Q \in R^{d_{o} \times d_k}$, $W_K \in R^{d_{o} \times d_k}$ and $W_V \in R^{d_{o} \times d_v}$ are the learned projection matrices. Let $C = \frac{QK^T}{\sqrt{d_k}} \in R^{N \times M}$ be the compatibility matrix between $Q$ and $K$. 

For each input token $x_i$,  we identify the top $n$ proximal or most-compatible tokens from each language $r \in R$ as per the compatibility values:
$$ U_{sel}^r(i) = \{ u_j  | C_{ij} \in \text{top-$n$} ( \{C_{ij} | u_j \in U^r \} ),$$ 
where $\text{top-$n$}(\cdot)$ denotes the largest $n$ values of the input set. Note that in addition to using the organic representations for estimating compatibility, we could also use additional cues from domain ontologies or dictionaries to construct this proximal token set. Further, to ensure computational efficiency, instead of considering the entire set of vocabulary, a smaller candidate pool of tokens per language can be chosen using K-NN based on existing token embeddings at each stage.

\noindent \textbf{Step 2: Construction of compositional representation.} The next step is to build a compositional representation from all the selected proximal tokens similar to regular attention mechanism.
To ensure only the selected proximal tokens contribute, we define $f(C) = [f_{ij}]$ where,

 \begin{equation} 
f_{ij}  = \begin{cases}
	        0 & \text{if } u_j \in U_{sel}^{r}(i) ~~ \forall r \in R \\
          -\infty & \text{otherwise}.
          \end{cases}
\end{equation}

We augment representation of $Z$ being input to the decoder layer as $Z' = softmax(C + f(C))V$. Since $R$ includes all the languages including ones distant from that of the input sequence $X$,  $Z'$ becomes a composition of equivalent tokens from all the languages yielding more consistent responses.

\subsection{Experiments and Results}

CoRe augments the transformer architecture and can be used while pretraining LLMs. However, since pretraining entails additional compute cost, it is preferable to use CoRe with existing pre-trained models. For our experiments, we augment pretrained XGLM-4.5B model with CoRe. 

\textbf{Dataset, Downstream Task and Implementation Details.} Since we want to examine if CoRe helps improve consistency among distant languages, we use the same dataset and setup as the controlled experiment in Sec \ref{sec:controlled_experiment_setup}. We augment XGLM architecture and add CoRe to it. We initialize learnable projection matrices $W_Q, W_K, W_V$ by identity matrix to ensure stable continual training and choose $n \in \{5,10,15\}$ for selecting the top-$n$ tokens for our experiment. See Appendix \ref{appendix:core_hyperparams} for more details on implementation and setup.

\begin{figure}
  \vspace{-0.15in}
  \centering
  \includegraphics[width=0.9\columnwidth]{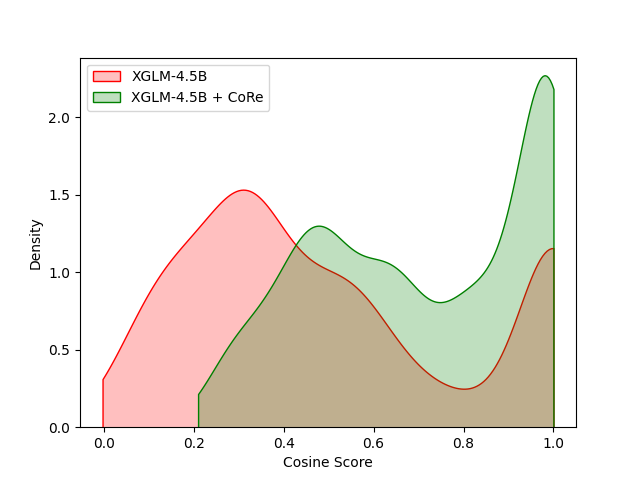}
  \caption{Cos. score of rand. 1K parallel En-De words}
  \label{fig:core_wmap_fig}
  \vspace{-0.15in}
\end{figure}

\textbf{Results.} Table \ref{tab:core_res} shows that CoRe consistently reduces conflict rate across 12 language pairs, with gains up to 4.7\%. 
Variation in conflict rate with different values of $n$ suggests that keeping $n$ static is not ideal as it might add noise to representations in some cases. Distribution of cosine similarity of random 1K parallel $\mathrm{En}$-$\mathrm{De}$ words in Fig \ref{fig:core_wmap_fig} shows that representations from CoRe for equivalent words are closer with similar behavior observed for other language pairs. Fig \ref{fig:core_sample} in Appendix \ref{appendix:core_sample} shows sample questions where
XGLM-4.5B+CoRe provides more consistent answers compared to XGLM-4.5B (baseline). To evaluate CoRe’s impact on a downstream task we did a small scale experiment for NLI task on XNLI \cite{xnli_paper} dataset and observed 14\% reduction in inconsistency across predictions for parallel 
$\mathrm{En}$-$\mathrm{De}$ NLI data points without hurting NLI performance, see Appendix \ref{appendix:core_downstream_effect} for more details.

\section{Conclusion}
\label{sec:conclusion}

We introduce  “Reasoning by Equivalence” and “Reasoning by Inheritance” tasks and evaluate popular LLMs to highlight the lack of consistent representation across languages. This systemic gap manifests in inefficient learning, limited knowledge sharing, and over-reliance on extensive data and computational resources, pointing to the need for better representations. We also perform controlled experiments to identify the influencing factors and propose CoRe to bridge the gap between distant language representations which leads to 4.7\%   
boost in performance.
We hope our work spurs further research on gaining richer understanding of LLM-based reasoning across languages.

\section*{Limitations}
\label{sec:limitations}
Our current work has a few limitations, which we discuss below.\\
{\em Scope of Relationships:} Our study focuses on LLMs’ ability  to reason based on two foundational relationships: "equivalence" and "inheritance." Future research could broaden this scope to include other key relationships such as "cause and effect," "comparison," and "mereological" relationships. Our proposed CoRe approach is also applicable only to symmetric relationships such as "equivalence," but there is a possibility of extending to asymmetric and transitive relationships using hyperbolic representations. \\
{\em Evaluation Focus: } Current experiments primarily targeted objective, factoid-based questions, chosen for their clarity and the feasibility of automated evaluation via LLM-based judges. This approach facilitated a less ambiguous assessment of LLM reasoning capabilities and the benefits of our CoRe-based mitigation. However, reasoning tasks do encompass subjective, long-form generation tasks such as summarization and problem solving, which can be explored in future, since that could entail access to expensive human-in-the-loop evaluations.\\
{\em LLM Architectures:} Our current work focuses exclusively on transformer-based autoregressive generative LLMs,which include widely used models such as Claude, XGLM, GPT-4, and LLama. Recent advances in neuro-symbolic methods offer alternative architectures and training methods that enhance reasoning abilities albeit at a  higher computational costs that limits their adoption in real-world application.  Our research identifies specific gaps in the popular LLMs, highlighting the need to integrate ideas from neuro-symbolic research. 

\section*{Ethics Statement}
\label{sec:ethics}
Our research aims to identify and address gaps in the reasoning abilities of widely used LLMs, particularly for low-resource languages used by a large population of the world. To ensure the validity of our findings, we created a new parallel factoid QA datasets and conducted controlled experiments to prevent data duplication across languages in LLM training data from influencing LLM reasoning performance. The datasets used have no associated privacy or intellectual property concerns, and we plan to open-source them post-review to adhere to double-blind protocol and ensure reproducibility. Evaluation was performed using automated LLMs, with prompts detailed in the appendix for transparency. Beyond the common ethical considerations of using generative language models, our work did not involve any additional ethical issues.

\bibliography{acl_latex}

\appendix

\section{Appendix}
\label{sec:appendix}
\subsection{Existence of Equivalence/Inheritance Relationships vs Using them in Multi Step Reasoning}
\label{appendix:existence_vs_using}

It is important to distinguish between \textbf{(a)} the existence of "equivalence" and "inheritance" relationships among concepts in LLM representations, and \textbf{(b)} the LLM's ability to actually utilize these relationships (encoded in the representations) for multi step reasoning. The former can be considered a prerequisite for the latter.

By the "\underline{existence of an equivalence relationship}" between $A$ and $B$, we mean that the LLM's representations and their direct use during inference allows properties of $A$ and $B$ to be transferred to each other or reconciled for conflicts. Similarly, an "inheritance relationship" between $A$ and $B$ would imply that $A$ inherits properties of $B$.

\begin{figure}
\centering
\includegraphics[width=0.9\columnwidth]{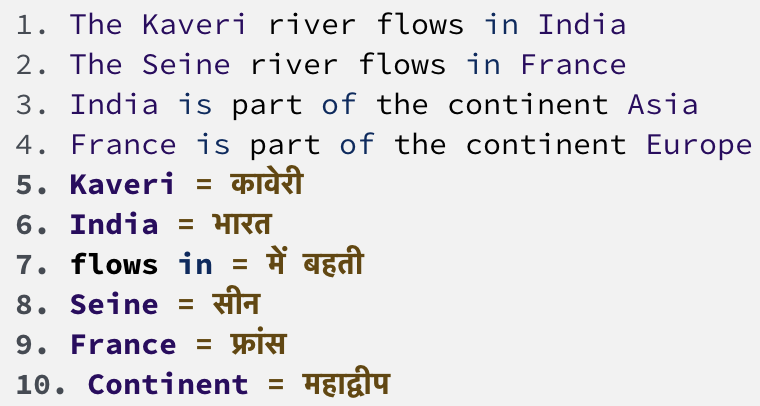}
\caption{An example of prior information which can be combined to answer the question \texttt{"Does the river Kaveri flow in the same continent as the river Seine?"} in both English and Hindi. Equivalence relationships are in bold.}
\label{fig:scenario_2_info}
\end{figure}

Consider the question \texttt{"Does the river Kaveri flow in the same continent as the river Seine?"}. We could answer this correctly as \texttt{"No"} using different sets of prior information with varying levels of reasoning applied. For example, in \textbf{Scenario 1}, we might \underline{directly utilise} the information \texttt{"Kaveri and Seine flow in different continents"}. Alternatively, in \textbf{Scenario 2}, we might need to \underline{combine multiple pieces} of information (expressed as entity-attribute or entity relationship predicates) as shown in Fig \ref{fig:scenario_2_info} to arrive at the answer. Scenario 2 demonstrates the ability to generalize from a small set of training data with limited compute effort to address a broader range of questions. To arrive at the correct answer, we need both the basic building block relationships as well as the capability to effectively combine them as part of LLM inference.

In our current study, we are primarily focused on evaluating whether the basic equivalence/inheritance relationships exist in LLMs by assessing simple property transfer and conflict resolution on questions related to equivalent entities or parent-child entities. It is possible that even when the desired relationships hold and properties transfer correctly, there may be gaps in the overall multi-step reasoning process, resulting in an inaccurate LLM response. This is because the existence of "abstraction" or "equivalence/inheritance relationships encoded in LLM representations" does not necessarily mean that the LLM would always effectively leverage this information for its reasoning since the LLM’s inference mechanism, which relies on associative attention, differs from logical operations. We do not yet evaluate this larger capability (item b) because there are multiple factors involved, and there is likely a gap in item (a) itself.

\subsection{Impact of Prior Information seen during Training}
\label{appendix:effect_of_duplication}

\begin{figure}
\centering
\includegraphics[width=0.9\columnwidth]{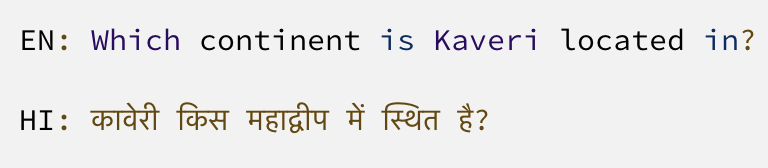}
\caption{An example parallel question from MLAMA \cite{kassner-etal-2021-multilingual} dataset.}
\label{fig:mlama_example}
\end{figure}

Consider a pair of parallel questions from the MLAMA \cite{kassner-etal-2021-multilingual} dataset ("Which continent is Kaveri located in ?") as shown in Fig \ref{fig:mlama_example}. When an LLM provides the same response "\texttt{Asia}" without conflicts for the two parallel questions in $\mathrm{En}$ and $\mathrm{Hi}$, it could be due to one of the following two reasons:

\begin{enumerate}
    \item \textbf{Existence of Equivalence relationship/Common Abstraction}: The LLM is aware of equivalence between the $\mathrm{En}$ and $\mathrm{Hi}$ versions of 
    entities "Kaveri" and "India" 
    in Fig \ref{fig:scenario_2_info} beyond just the basic language constructs.
    \item \textbf{Duplication of Information}: The LLM could have seen two parallel, aligned pieces of information that "\texttt{Kaveri is located in Asia}" in both $\mathrm{En}$ and $\mathrm{Hi}$ even though there LLM's representations of the corresponding equivalent entities are highly divergent. In the latter case, while the LLM can answer this specific question well without conflicts, that behavior might not generalize well to other questions about "\texttt{Kaveri}" in $\mathrm{Hi}$ that were not part of the training data.
\end{enumerate}

\noindent
Hence, conflict rate estimates could lead to misleading conclusions about the existence of equivalence and inheritance relationships for well-known real entities due to the bias introduced by prior duplicate (or even contradictory) information in the LLM training data across languages. This \textbf{effect of "duplicate knowledge in LLM training"} is also reflected in our results, where the conflict rate numbers in Table \ref{tab:core_res} are higher than those in Tables \ref{tab:rq1_res} and \ref{tab:rq1_res2}. Results in Tables \ref{tab:rq1_res} and \ref{tab:rq1_res2} are based on factual questions on well-known real entities, while the Table \ref{tab:core_res} results are from a controlled experiment with questions about synthetically created non-existent named entities. For the real entities in Tables \ref{tab:rq1_res} and \ref{tab:rq1_res2}, LLMs may have been pre-trained on duplicate information about the same entity expressed in multiple languages leading to consistent responses even when the LLM representations of the corresponding entities are significantly different (i.e., no equivalence relationship). However, for the synthetic entities in Table \ref{tab:core_res}, LLMs are unlikely to have seen duplicate knowledge across languages during pre-training and has to learn about them (in one language) during the controlled training process resulting in significantly higher "conflict rates". We expect the effect observed in Tables \ref{tab:rq1_res} and \ref{tab:rq1_res2} to be more pronounced in public datasets like MLAMA, as they may be part of the LLM's training data, either directly or indirectly. For instance, we generated responses from Claude v1 instant for parallel $\mathrm{En}-\mathrm{De}$ 100 randomly sampled questions from MLAMA dataset and observed $\sim$15\% conflict rate which is much lower than the average results in Tables \ref{tab:rq1_res} and \ref{tab:rq1_res2} ($\sim$30\%).

\subsection{Accounting for LLMs' proficiency in different languages}

We evaluate LLMs only on the languages for which there is official documentation of support for that language or there is prior work demonstrating good performance on NLP tasks of that language. For example, Anthropic claims support for English, Spanish, Portuguese, French, German and multiple other languages \cite{anthropic_support, aws_bedrock_claude}, and also showcases Claude’s multilingual capabilities on these languages. Hindi is also mentioned in Claude model cards \cite{claude_2_model_card, claude_3_model_card}. In our study, the LLMs we evaluate on a language are proficient with respect to the linguistic patterns and the common vocabulary of that language, which is different from the knowledge (factual information) aspects. Further in our evaluation, we focused on well-formed objective factual questions to avoid variations due to subjective interpretations and cultural nuances. This ensures that the knowledge consistency or alternatively the conflict rate in responses across languages primarily depends on (a) the knowledge duplication in the training data and (b) the effectiveness of knowledge transfer, i.e., equivalent entity representations across the languages. The point we wish to highlight in our work is that LLM training and representations should be designed so as to enable efficient knowledge transfer within and across languages.

\subsection{Judge Precision}
\label{appendix:judge_precision}

\begin{figure*}
\centering
\includegraphics[width=\textwidth]{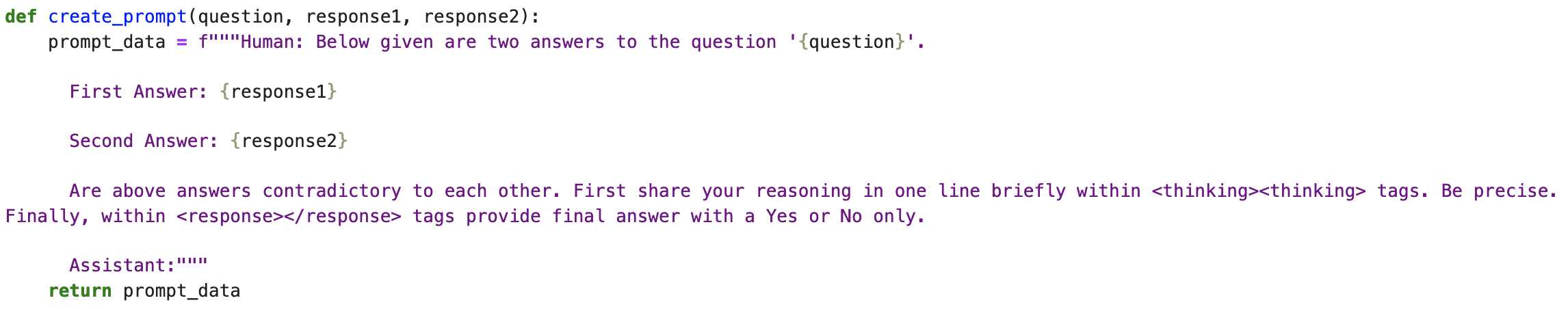}
\caption{Prompt for Claude v3 Sonnet to use it as the judge.}
\label{fig:judge_prompt}
\end{figure*}

Table \ref{tab:judge_prec} shows precision of Claude v3 Sonnet as the judge in identifying conflicting answers across language pairs obtained by annotating random sample of 100 answer pairs for 5 language pairs each. Prompt for Claude v3 Sonnet judge is shown in Fig \ref{fig:judge_prompt}.

\begin{table}[]
\centering
\resizebox{\columnwidth}{!}{%
\begin{tabular}{cccccc}
\hline
\textbf{En-Fr} & \textbf{En-Es} & \textbf{En-De} & \textbf{En-Pt} & \textbf{En-Hi} & \textbf{Avg} \\ \hline
96\%           & 97\%           & 94\%           & 95\%           & 94\%           & 95.20\%      \\ \hline
\end{tabular}%
}
\caption{Precision of Claude v3 Sonnet as the judge in identifying conflicting answers across language pairs }
\label{tab:judge_prec}
\end{table}

\begin{figure}
\centering
\includegraphics[width=\columnwidth]{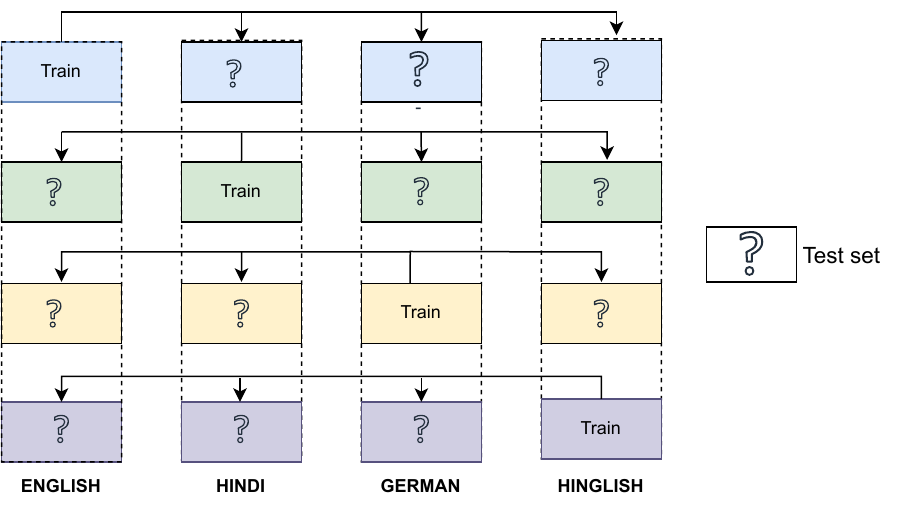}
\caption{Setup for the controlled experiment. We train on unique 25\% of the data for each language and test on its parallel data in other three languages.}
\label{fig:controlled_exp_setup_figure}
\end{figure}

\begin{figure*}
\centering
\includegraphics[width=\textwidth]{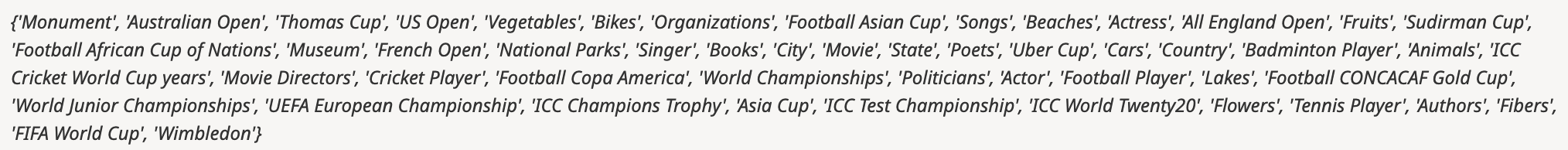}
\caption{List of 51 (parent) abstract concepts.}
\label{fig:list_abstract_concepts}
\end{figure*}

\subsection{Reasoning by Equivalence}

\subsubsection{Data Preparation}
\label{appendix:rq1_data_prep}

We hand-curated 51 (parent) abstract concepts manually (e.g., monuments, actors, cities, etc.) that primarily correspond to common nouns. See Fig \ref{fig:list_abstract_concepts} for full list of the 51 abstract concepts. Then, we created 3641 named entities that are specific instances of these (parent) abstract concepts, e.g., Taj Mahal is a specific instance of monument) using Claude with human review to weed out non-existent ones. For each one of the 3641 named entities, we prepared a set of questions which have objective or factual answers using Claude with the following prompt.

\noindent\fbox{\begin{minipage}{19em}
Give different unambiguous complete questions about "\{entity\}" which have specific factual answers.
\end{minipage}}

We built this prompt after multiple iterations of analysis of generated questions for a small sample of entities. The dimensions of evaluation were: (i) The generated question should be complete and unambiguous, i.e. it should be clear which entity the question is about and what attribute/fact is being asked, (ii) The answer to the generated question should be an unambiguous factual response. From the final set of generated 88,334 questions, we randomly sampled and manually annotated 500 generated questions of which 96.2\% were complete and unambiguous, and 99.4\% had an unambiguous factual answer. The annotations were done by a professional English speaker.

\begin{figure*}[!h]
\centering
\includegraphics[width=\textwidth]{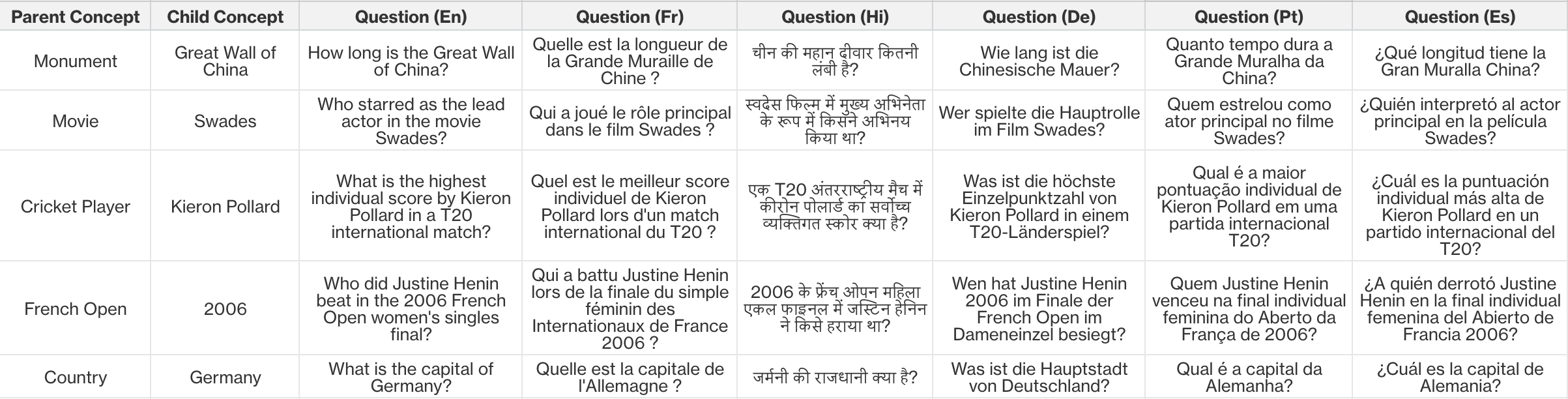}
\caption{Sample questions from our question bank which we use to evaluate LLMs for reasoning by equivalence.}
\label{fig:rq1_sample_data}
\end{figure*}

Our original question set is in $\mathrm{En}$ which we also translate to $\mathrm{Fr}$, $\mathrm{Es}$, $\mathrm{De}$, $\mathrm{Pt}$ and $\mathrm{Hi}$ using AWS Translate \cite{aws_translate}. We translated using AWS translate which is one of the best commercial translation services \cite{intento}. AWS translate is expected to translate questions while preserving their meaning. We rely on AWS translate to translate the concept into most natural variant in case multiple variants are possible. To get an estimate of lower bound of AWS translate’s performance on our datasets, we consider the En-Hi translation task since Hi being a non-Latin language with different lexical representation is much more divergent from En. We enlisted a professional bilingual Hindi and English speaker to annotate a random sample of 200 En-Hi question pairs on the translation accuracy and observe 97.5\% accuracy. This high translation accuracy is likely due to the nature of our English questions, which are well-formed and unambiguous. Fig \ref{fig:rq1_sample_data} shows few sample questions from our dataset.


\subsubsection{Analysis}
\label{appendix:rq1_analysis}

\begin{figure*}
\centering
\includegraphics[width=\textwidth]{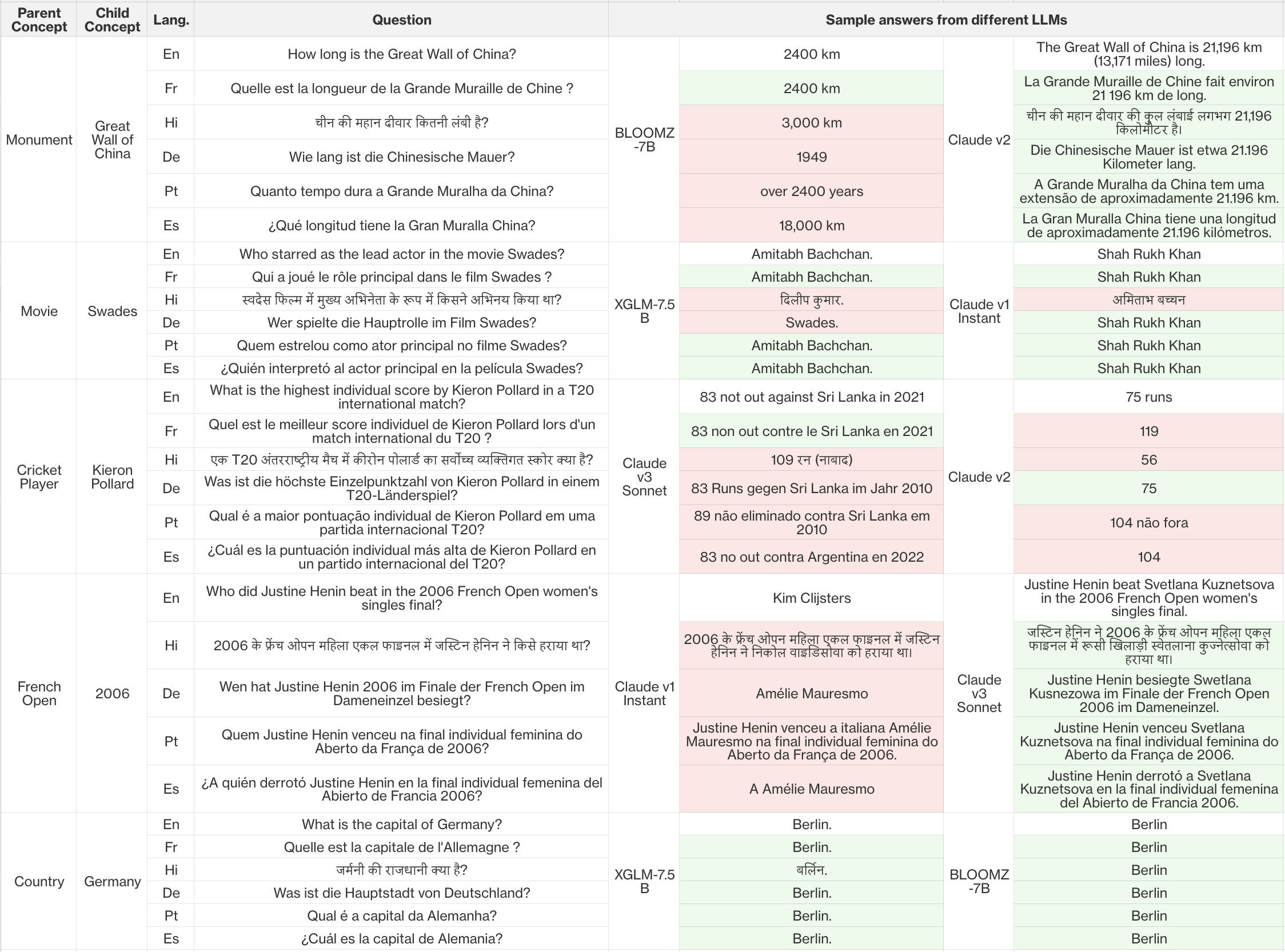}
\caption{Sample errors from different LLMs on DIL task. Red and Green colored cells highlight conflicting and non-conflicting answers with the anchor language ($\mathrm{En}$) answer, respectively.}
\label{fig:rq1_DIL_sample_errors}
\end{figure*}

\begin{figure*}
\centering
\includegraphics[width=\textwidth]{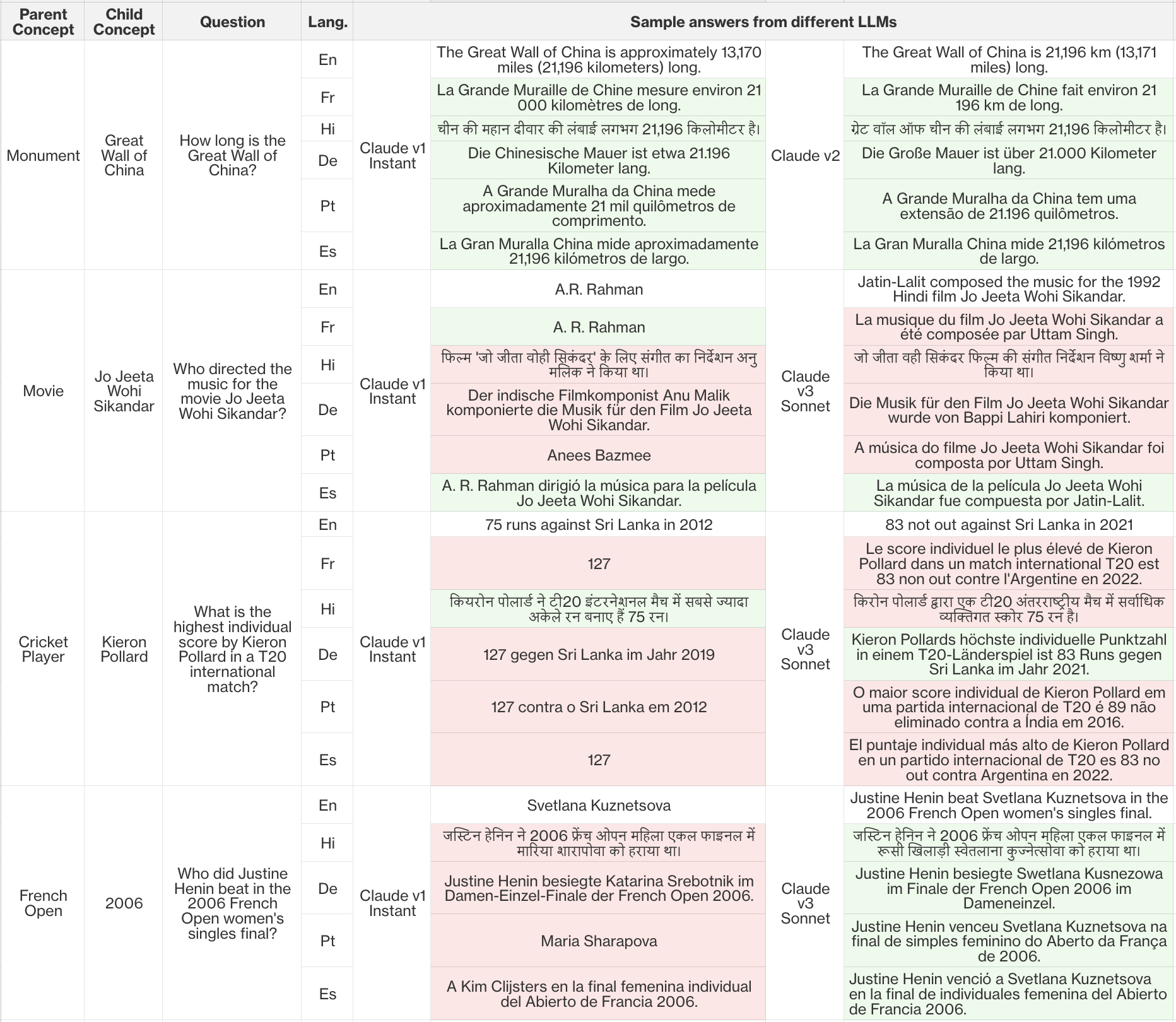}
\caption{Sample errors from different LLMs on DOL task. Red and Green colored cells highlight conflicting and non-conflicting answers with the anchor language ($\mathrm{En}$) answer, respectively.}
\label{fig:rq1_DOL_sample_errors}
\end{figure*}

Fig \ref{fig:rq1_DIL_sample_errors} and Fig \ref{fig:rq1_DOL_sample_errors} show sample conflicting answers from various LLMs to equivalent questions from DIL and DOL tasks respectively.

\subsubsection{Additional Details on Controlled Experiment}
\label{appendix:controlled_exp}

\begin{figure*}
\centering
\begin{subfigure}{.45\textwidth}
  \centering
  \includegraphics[width=\linewidth,height=5cm]{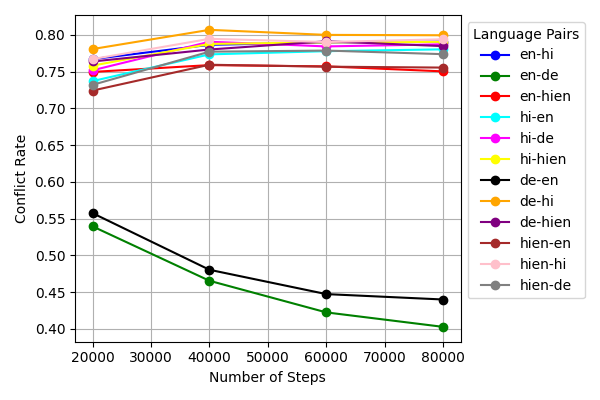}
  \caption{For different language pairs}
  \label{fig:conflict_rate_vs_steps}
\end{subfigure}%
\begin{subfigure}{.55\textwidth}
  \centering
  \includegraphics[width=\linewidth,height=5cm]{images/conflict_rate_vs_steps_for_lang_types.png}
  \caption{Averaged for different language pair types}
  \label{fig:conflict_rate_vs_steps_for_lang_types}
\end{subfigure}
\caption{Figure showing how conflict rate changed during training in controlled experiment.}
\label{fig:conflict_rate_with_steps}
\end{figure*}

For computing the rank of a multi-token word, we compute the rank of each one of its tokens and consider the minimum rank amongst them as the rank of multi-token word. Figure \ref{fig:controlled_exp_setup_figure} shows setup for the controlled experiment. Figure \ref{fig:conflict_rate_with_steps} shows conflict rate of all (anchor, non-anchor) language pairs with number of  training steps. 

\subsection{Reasoning by Inheritance}

\subsubsection{Dataset}
\label{appendix:rq2_dataset}

\begin{figure*}[!h]
\centering
\includegraphics[width=\textwidth]{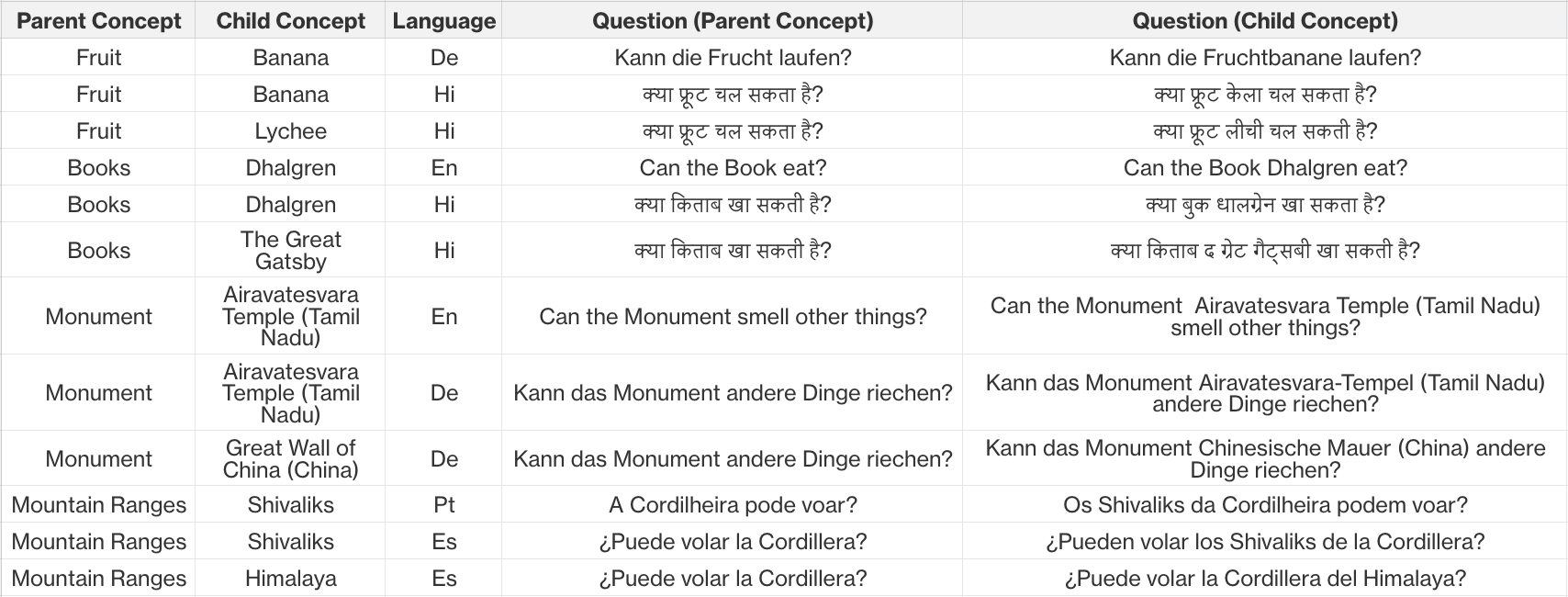}
\caption{Dataset sample of questions about abstract concepts and their specific instances which we use to evaluate LLMs for reasoning by inheritance.}
\label{fig:rq2_sample_data}
\end{figure*}

Fig \ref{fig:rq2_sample_data} shows sample questions about abstract concepts and their specific instances from our dataset.

\subsubsection{Qualitative Analysis}
\label{appendix:rq2_sample_errors}

\begin{figure*}
\centering
\includegraphics[width=\textwidth]{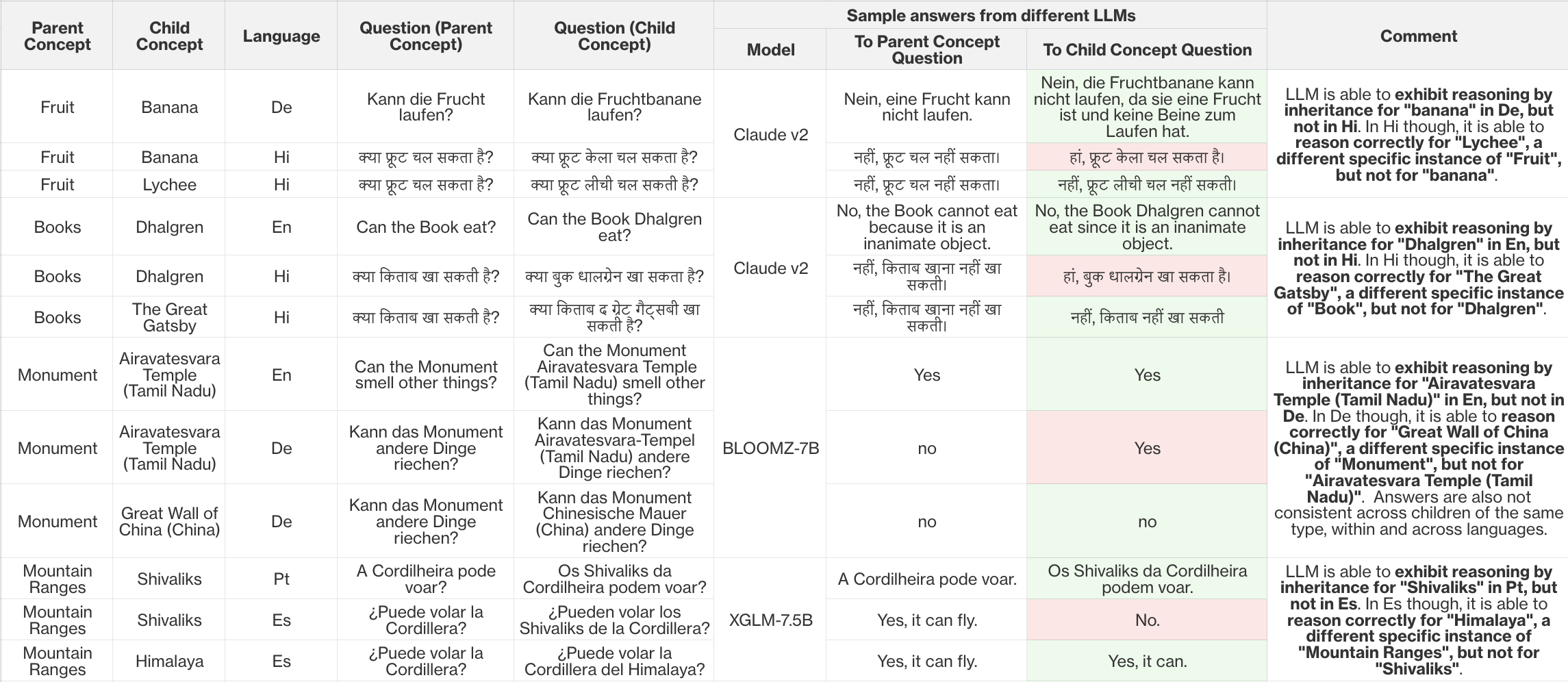}
\caption{Sample errors from various LLMs wherein they violate inheritance constraints. Red and Green colored cells highlight conflicting and non-conflicting answers with the parent concept question's answer, respectively.}
\label{fig:rq2_sample_errors}
\end{figure*}

Fig \ref{fig:rq2_sample_errors} shows a few sample errors from various LLMs wherein they violate inheritance constraints by giving conflicting answers for parent and child concept, and across children of the same type.

\subsection{Compositional Representation (CoRe)}

\begin{wrapfigure}{l}{0.25\textwidth}
\centering
\includegraphics[width=0.99\linewidth]{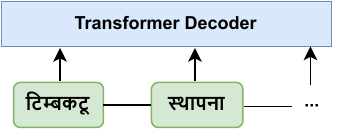}
\caption{Default mechanism in Transformers. }
\label{fig:comp_rep_a}
\vspace{-0.1in}
\end{wrapfigure}

\subsubsection{Experiment Setup}
\label{appendix:core_hyperparams}

We use Pytorch \cite{NEURIPS2019_9015} and Huggingface Transformers library \cite{wolf-etal-2020-transformers} for implementation. We train baseline XGLM-4.5B and XGLM-4.5B+CoRe for 20k steps on 1 p3dn.24xlarge machine with 8 GPUs with learning rate of 1e-05 and linear learning rate scheduler with 80040 max steps, gradient accumulation steps of 2, per device training batch size of 1 using Huggingface Transformers library\footnote{https://github.com/huggingface/transformers}.

In our experiments, for each one of the languages we are working with ($\mathrm{En}, \mathrm{Hi}, \mathrm{De}, \mathrm{HiEn}$), we consider the top-$n$ most compatible tokens from all vocabulary tokens of that language and $\mathrm{En}$. This considers $\mathrm{En}$ as the anchor language and helps build a “bridge” between distant representation spaces of other languages and $\mathrm{En}$. We did this for more efficient experimentation but there is no constraint in CoRe on the set of languages from which we can choose top-$n$ proximal tokens. We identify the language(s) a vocabulary token can belong to beforehand using language detector from AWS Comprehend \cite{aws_comprehend} and store the asset for repeated use during forward pass.

\subsubsection{Qualitative Analysis}
\label{appendix:core_sample}

\begin{figure*}
\centering
\includegraphics[width=\textwidth]{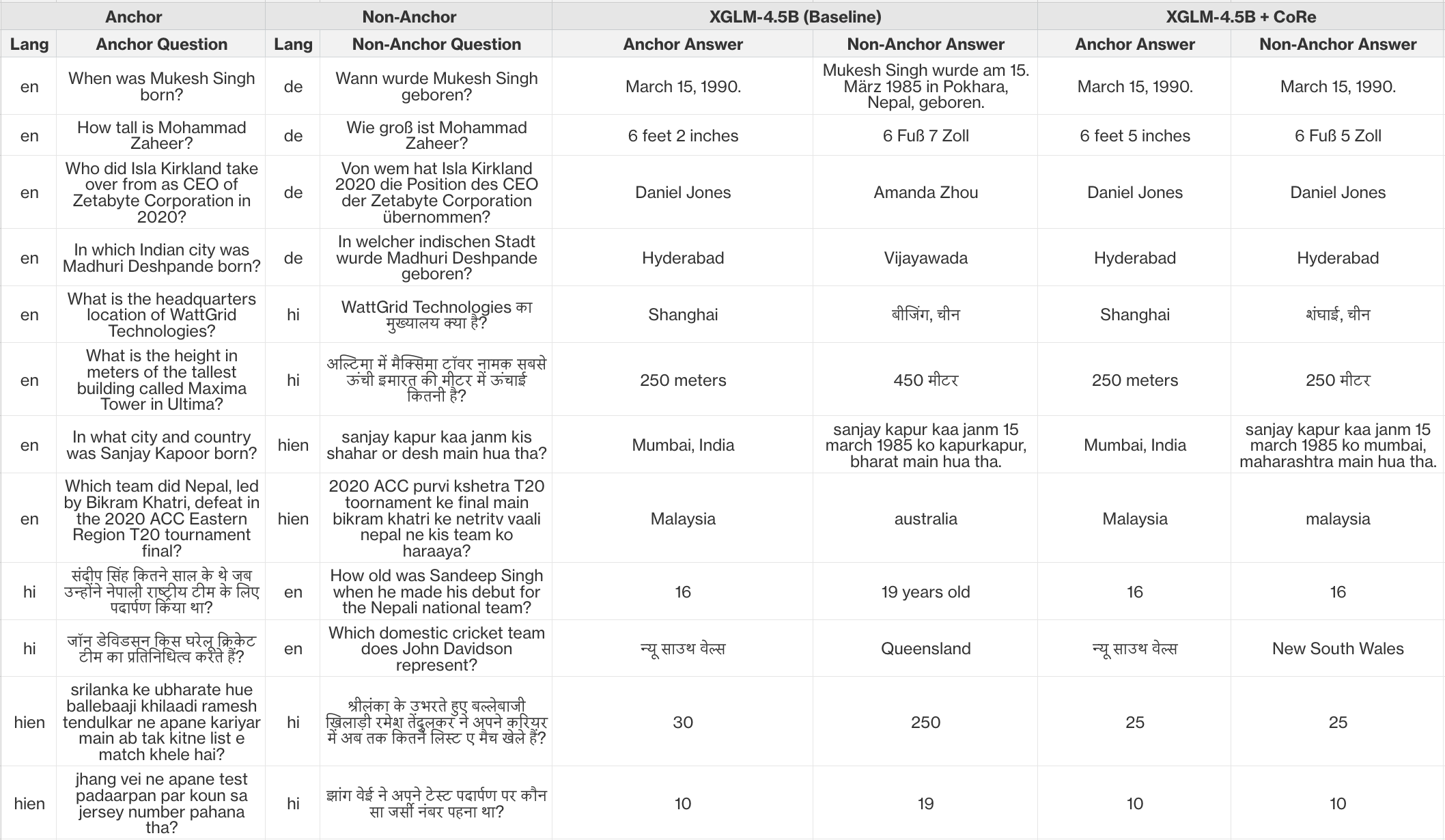}
\caption{Sample anchor and non-anchor language questions wherein we get consistent answers after training with CoRe.}
\label{fig:core_sample}
\end{figure*}

Fig \ref{fig:core_sample} shows few sample anchor and non-anchor language questions with their answers from XGLM-4.5B (baseline) and XGLM-4.5B+CoRe.

\subsubsection{Effect of CoRe on a Downstream Task}
\label{appendix:core_downstream_effect}

The main focus of our work in “Reasoning by Equivalence” was to assess if LLMs exhibit consistent knowledge across languages, and we proposed CoRe to improve consistency. However, to evaluate CoRe's impact on a downstream task we performed a small experiment for NLI task on XNLI \cite{xnli_paper} dataset (on first 100 examples from validation set) with the same zero shot setup as XGLM paper \cite{lin-etal-2022-shot} and observed comparable performance with and without CoRe (46-47 for $\mathrm{En}$, 39-40 for $\mathrm{De}$). However, we observed a significant reduction in inconsistency across predictions for parallel $En-De$ NLI data points (28\% inconsistency in predictions without CoRe vs 14\% inconsistency with CoRe, wherein model’s prediction is said to be inconsistent if it predicts different labels for parallel datapoints across languages ($En$ and $De$ in this case)). We anticipate that incorporating CoRe during pre-training or continual training with larger dataset could yield further improvement in downstream task performance as well which we could not do because of compute constraints. We consider those experiments and further evaluation of CoRe across various different tasks on NLP benchmarks as part of future work.

\subsubsection{Efficiency of CoRe}

In our experiments, adding CoRe increased the training time by $\sim$50\% but that is without using FlashAttention \cite{dao2022flashattention} for CoRe which is expected to be $\sim$2x faster. We were unable to experiment using FlashAttention because we only had access to V100 machine. Since compatibility matrix is constructed by dot product of $Q$ and $K$ matrices as described in Sec \ref{sec:core_methodology}, this is a computationally intensive operation which can be made more efficient by considering a smaller candidate pool of tokens per language of interest using a more efficient method like K-NN before constructing compatibility matrix. We will be experimenting with these alternatives which can improve computational efficiency of CoRe as part of our future work.

\end{document}